# Ensemble Detection of Single & Multiple Events at Sentence-Level

Luís Marujo, LTI/CMU, USA and INESC-ID/IST, Portugal
Anatole Gershman, LTI/CMU, USA
Jaime Carbonell, LTI/CMU, USA
João P. Neto, INESC-ID/IST, Portugal
David Martins de Matos, INESC-ID/IST, Portugal

## INTRODUCTION

Modern Newspapers have been organized into news articles. This structure was kept since their invention in the early 17th century. These articles are usually organized in an "inverted pyramid" structure, placing the most essential, novel and interesting elements of a story in the beginning and the supporting materials and secondary details afterwards. This structure was designed for a world where most readers would read one newspaper per day and one article on a particular subject. This model is less suited for today's world of online news where readers have access to thousands of news sources. While the same high-level facts of a story may be covered by all sources in the first few paragraphs, there are often many important differences in the details "buried" further down. Readers who are interested in these details have to read through the same materials multiple times. The problem becomes even more pronounced when difference sources of information, such as Broadcast News, Blogs, and News sites, are taken into account.

Automated identification of event instances can help finding specific content of interest to the user. Event detection is a fundamental Information Extraction task, which has been explored largely in the context of QA, TDT and Summarization. More recent work using the TDT datasets on Event Threading [8] tried to organize news articles about armed clashes into a sequence of events, but still assumed that each article described a single event. Passage Threading [8] extends the event threading by relaxing the one event per news article assumption and it was restricted to the identification of "violent" paragraphs by a binary classifier. In this paper, we explore both multi-label and multiclass techniques to identify the sentences in a document that contain specific event types. First, we replicated and enhanced the work described in Naughton's work [3]. We used a meta-Machine Learning algorithm and added several new features, achieving 4.6% improvement in F1 scores over the results reported by [3]. This initial scenario is a specific case of a single-label (multi-class) classification problem, where each sentence in a document is classified to one and only one target event or no event. The result is a simplification of the problem given the low occurrence of multi-event sentences in the corpus used.

We use the ACE 2005 Multilingual Corpus [6] in our experiments. Although it was annotated for 33 different event types, only a few event types have enough instances to train a classifier. Consequently, we only selected 6 event types. Because a single sentence may include several types of events, e.g., sentence "Two people were killed in a terrorist attack on a train" contains three types of events: "die", "attack", and "transport", we explored supervised classification methods that capture the relationships between event types, such as the Classifier Chains method and its ensemble. To make comparison between different classifiers meaningful, we need to use balanced test set. Given the absence of publicly available balanced dataset, we created EVNE2013 dataset with 100 news articles covering 10 violent and economic event types.

The main contributions of this work are: (1) Creation of EVNE2013 dataset; (2) Reduction of the hard  imbalanced multi-label classification problem with low number of occurrences of multiple labels per instance to an more tractable imbalanced multiclass problem with better results; (3) We report the results of adding new features, such as sentiment strength, rhetorical signals, domain-id (source-id and date), and key-phrases in both single-label and multi-label event classification scenarios; (4) Application of multi-label classification methods.

**CORPUS**

The ACE2005 Corpus was created for the ACE evaluations, where an event is defined as a specific occurrence involving participants. The corpus has 33 event types. From these 33 events, only the following 6 have a high number of instances or sentences in the corpus: Die, Attack, Transport, Meet, Injure, and Charge-Indict. About 16% of the sentences contain at least 1 event. From those sentences, 15% of the sentences are classified as multi-event (or multi-label), for instance the sentence "3 people died and 2 injured when their vehicle was attacked" involves 4 event types (or one event with 4 event type labels). These multi-event sentences correspond to only 2.50% of the corpus.

Imbalanced datasets, such as ACE2005 Corpus, are known to be a more realistic scenario for event recognition than balanced datasets. However, for the purpose of evaluation, we created a selection of sentences that form a balanced set: the EVNE2013 dataset[1], with 100 news articles covering 10 violent and economic event types: Armed Clashes, Bankruptcy, Change of CEO, Legal Trouble, Mergers, Sex abuse, Street protest, Strike, Suicide Bombing, and Terrorism Bombing. The news articles were collected from 63 sources.

**FEATURE EXTRACTION**

The spoken transcripts documents found in the ACE2005 corpus contain raw ASR single-case words with punctuation. This means that the transcriptions were either manually produced or were generated by a standard ASR with minimal manual post-processing. Absence of capitalization is known to negatively influence the performance of parsing, sentence boundaries identification, and NLP tasks in general. Recovery of capitalization entails determining the proper capitalization of words from the context. This task was performed using a discriminative approach described in [1]. We capitalized every first letter of a word after a full stop, exclamation, and question mark. After true-casing, we automatically populate three lists for each article: list K of key phrases, list V of verbs, and list E of named entities. The key phrase extraction is performed using a supervised automatic key phrase extraction method [2]. Verbs are identified using Stanford Parser, and named-entities using Stanford NER. This extraction is performed over all English documents of the corpus. The K, V, and E lists are used in the extraction of lexical features and dependency parsing-based features. The lists K and V were also augmented using WordNet synsets to include less frequent synonyms. Furthermore, we manually created list M of modal verbs, and list N of negation terms.

The feature space for the classification of sentences consists of all entries in the lists V, E, K, M, and N which are corpus specific. The value of each feature is the number of its occurrence in the sentence. These numbers indicate the description of events by numbering the number of participants, actions, locations, and temporal information. We have also explored other uncommon types of features: Rhetorical Signals [2] and Sentiment Scores [5]. Finally, we removed all features with constant values across classes. This process reduced by half the number of features and improved the classification results.

**SINGLE-EVENT DETECTION**

A state-of-art way to solve a text multi-class problem, like single event detection, is to use SVM techniques [11]. Sometimes SVM can be improved using Adaboost.M1 particularly on imbalanced datasets [9]. Thus, we explored both SVM and Adaboost.M1+SVM classifiers.

**MULTI-EVENT DETECTION**

Multi-Event detection at sentence-level is a multi-label classification problem. As a simple baseline for comparison, we use Binary Relevance (BR) classification. The BR method is the best-known multi-label classification, which assumes L (number of labels) independent binary

---

[1] EVNE2013 is available at https://www.l2f.inesc-id.pt/wiki/index.php/EVNE2013

problems.

We have also investigated a relatively new multi-label technique Classifier Chain (CC) method because it performed better than the BR in several multi-label problems [4].

The CC method uses binary transformations as in the BR method. However, CC diverges from BR as each feature vector is augmented with the binary labels of all previous classifiers, thus creating a CC of binary classifiers. The classifier process begins at the first classifier $h_1$ and propagates the predictions along the chain. Such propagation is designed to capture correlation between labels overcoming BR's main limitation. But, at the same time, the chain also propagates errors. Such propagations can be problematic for long chains. As the Ensemble of Classifier Chain (ECC) reduces those effects by exploring several chain orders, it was also investigated in the multi-label problem.

**EVALUATION AND RESULTS**

| Labels | Ada.SVM (all f.) | SVM (all f.) | SVM (Base f.) | SVM (No S.-Id) | SVM (No Sent. A.) | SVM (No Rhet. S. | SVM (No Dep. Pars.) |
|---|---|---|---|---|---|---|---|
| N | **0.824** | 0.822 | 0.661 | 0.700 | 0.815 | 0.805 | 0.818 |
| Die | 0.691 | 0.681 | 0.620 | 0.643 | 0.691 | **0.693** | 0.683 |
| C.-Ind. | 0.709 | 0.709 | **0.761** | 0.706 | 0.692 | 0.725 | 0.735 |
| Transp. | 0.673 | 0.676 | 0.579 | 0.633 | 0.630 | **0.677** | 0.665 |
| Meet | **0.655** | 0.652 | 0.586 | 0.614 | 0.649 | 0.632 | 0.632 |
| Injure | 0.627 | 0.610 | **0.711** | 0.667 | 0.610 | 0.557 | 0.619 |
| Attack | **0.780** | 0.776 | 0.692 | 0.739 | 0.748 | 0.763 | 0.772 |
| Avg. | **0.708** | 0.704 | 0.659 | 0.671 | 0.691 | 0.693 | 0.703 |

*Table 1 - Single-Label F1 classification results in the ACE 2005.*

The inclusion of the AdaBoost.M1 in the ACE2005 dataset (Table 1), on average across all 6 labels, lifts the F1 scores by 0.6% when compared with SVM without AdaBoost.M1 and obtained the best global performance in the single-event task (70.8% and 65.2% in the EVNE2013 – Table 2). We have also investigated the influence of the new features introduced in this work by using all features but the ones under test and a SVM multi-class classifier. These new sets of features raised the classification results by 6.8%. The inclusion of the domain-Id features raised the average F1 scores by 5%, which is the highest contribution among the new features. It was followed by the sentiment analysis features with 2%. The relevance based features, such as the rhetorical features and dependency parse based features had the lowest contribution with respectively 1.6% and 0.1%. However, the contribution of the domain-id features to individual F1 value of "Injure" label is -5.7% and dependency parse based features contribute -0.9% to the identification "Injure". One of the possible explanations for these two exceptions is the imbalanced distribution of the event types, which bias the classifier towards more frequent event types.

| Labels | F1 |
|---|---|
| N | 0.761 |
| Armed Clashes | 0.720 |
| Bankruptcy | 0.831 |
| Change of CEO | 0.373 |
| Legal Trouble | 0.519 |
| Mergers | 0.614 |
| Sex abuse | 0.667 |
| Street protest | 0.734 |
| Strike | 0.538 |
| Suicide Bombing | 0.742 |
| Terrorism Bombing | 0.677 |
| Average | 0.652 |

*Table 2 - F1 classification results in EVNE2013 dataset*

When we compared the performance of the baseline in Table 3, we concluded that ECC outperformed it by 2.8%. Nevertheless, the multiclass baseline proposed in this work improves near 1.6% over ECC. The ECC method is only outperformed in 1 event type by the BR and CC. These corresponded to less frequent event type.

| **Labels** | **BR** | **CC** | **ECC** |
|---|---|---|---|
| N | 0.805 | 0.811 | **0.862** |
| Die | 0.729 | 0.711 | **0.745** |
| Charge-Ind. | **0.468** | **0.468** | 0.456 |
| Transport | 0.636 | 0.636 | **0.658** |
| Meet | 0.595 | 0.607 | **0.619** |
| Injure | 0.731 | 0.731 | **0.749** |
| Attack | 0.752 | 0.741 | **0.760** |
| Avg. | 0.673 | 0.672 | **0.692** |

*Table 3 Multi-Label results in the ACE 2005*

**CONCLUSIONS**

In this paper, we described the problem of classifying events at sentence level which benefits several NLP-based systems benefit from advances in this task such as, personalization, recommendation, question answering, and summarization. The classification of events at sentence-level is a clear multi-label, multi-class problem. However, the number of sentences belonging to more than one event type is frequently limited. Consequently, we experimented with a relaxation of the imbalanced multi-label classification problem by modeling it as multi-class classification problem. This simplification raised the classification results across 6 event types plus no-event in the ACE2005 dataset above the ECC that was significantly better than the CC and BR confirming the hypotheses of correlation between events and propagation of classification errors in the CC.

To further evaluate the performance of the classifiers without the bias of the imbalanced distribution and given that we could not find a balanced dataset, we created EVNE2013 dataset. Based on the two datasets, we found that the domain-id and date were particularly useful features. The sentiment analysis and rhetorical devices feature were found to have also significative effect.

In future work, we will investigate the effects of classifying a large number of event types by exploring a learning framework based on the inclusion of "nugget" information based on user feedback [10].

**8.REFERENCES**